\documentclass[runningheads]{llncs}

\usepackage[year=2024,ID=1617]{eccv}

\usepackage{eccvabbrv}

\usepackage{graphicx}
\usepackage{booktabs}
\usepackage[accsupp]{axessibility}  

\usepackage[pagebackref,breaklinks,colorlinks]{hyperref}

\usepackage{orcidlink}
\usepackage{multirow}

\begin{document}

\title{Task-adaptive Q-Face} 

\titlerunning{Q-Face}

\author{Haomiao Sun\inst{1,2} \and
Mingjie He\inst{1,2} \and
Shiguang Shan\inst{1,2} \and \\Hu Han\inst{1,2} \and Xilin Chen\inst{1,2}}

\authorrunning{H.~Sun et al.}

\institute{Key Lab of Intelligent Information Processing of Chinese Academy of Sciences (CAS), Institute of Computing Technology, CAS, Beijing, 100190, China \and University of Chinese Academy of Sciences, Beijing, 100049, China\\
\email{haomiao.sun@vipl.ict.ac.cn\\\{hemingjie, sgshan, hanhu, xlchen\}@ict.ac.cn}}



\maketitle

\begin{abstract}

Although face analysis has achieved remarkable improvements in the past few years, designing a multi-task face analysis model is still challenging. Most face analysis tasks are studied as separate problems and do not benefit from the synergy among related tasks. In this work, we propose a novel task-adaptive multi-task face analysis method named as Q-Face, which simultaneously performs multiple face analysis tasks with a unified model. We fuse the features from multiple layers of a large-scale pre-trained model so that the whole model can use both local and global facial information to support multiple tasks. Furthermore, we design a task-adaptive module that performs cross-attention between a set of query vectors and the fused multi-stage features and finally adaptively extracts desired features for each face analysis task. Extensive experiments show that our method can perform multiple tasks simultaneously and achieves state-of-the-art performance on face expression recognition, action unit detection, face attribute analysis, age estimation, and face pose estimation. Compared to conventional methods, our method opens up new possibilities for multi-task face analysis and shows the potential for both accuracy and efficiency.
\end{abstract}

\section{Introduction}
\label{sec:intro}
Face analysis is the process of understanding and interpreting human facial characters from images or videos. Typical face analysis tasks include face recognition \cite{liu2017sphereface, wang2018cosface, deng2019arcface, zhong2021face}, expression classification \cite{farzaneh2021facial, liu2022adaptive, mao2023poster, wasi2023arbex}, attribute recognition \cite{liu2015deep, mahbub2018segment, hand2017attributes, zhuang2018multi}. These face analysis tasks have achieved remarkable success in the past few years by developing deep learning methods. However, most face analysis tasks are studied as separate problems and do not benefit from the synergy among related tasks. Unlike machines, humans can perceive other people’s multiple facial characteristics, such as identity, emotion, age, and gender, with only one single glance. This is a reasonable instinct since these facial characters are not independent but interrelated and complementary. For example, recognizing a person’s identity can help infer their emotion. Similarly, recognizing a person’s age and gender can provide additional information about their identity and emotions. Thus, humans can achieve a more comprehensive and accurate understanding of others by integrating multiple facial characters rather than focusing on one aspect alone.

\begin{figure}[t]
 \centering
\subcaptionbox{Task-specific Methods\label{fig:TS}}
{\includegraphics[height=0.21\linewidth]{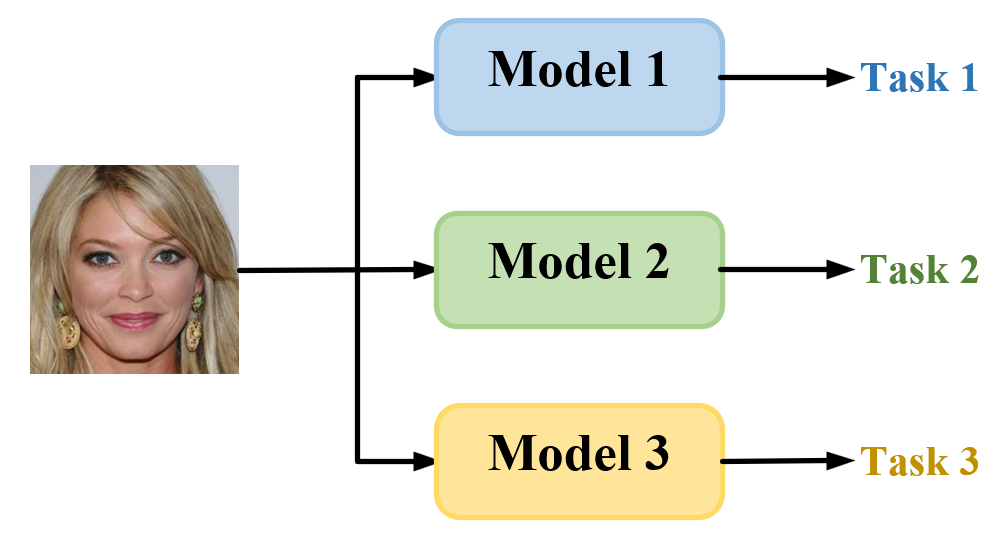}}
\subcaptionbox{Previous Multi-task Methods\label{fig:MH}}
{\includegraphics[height=0.21\linewidth]{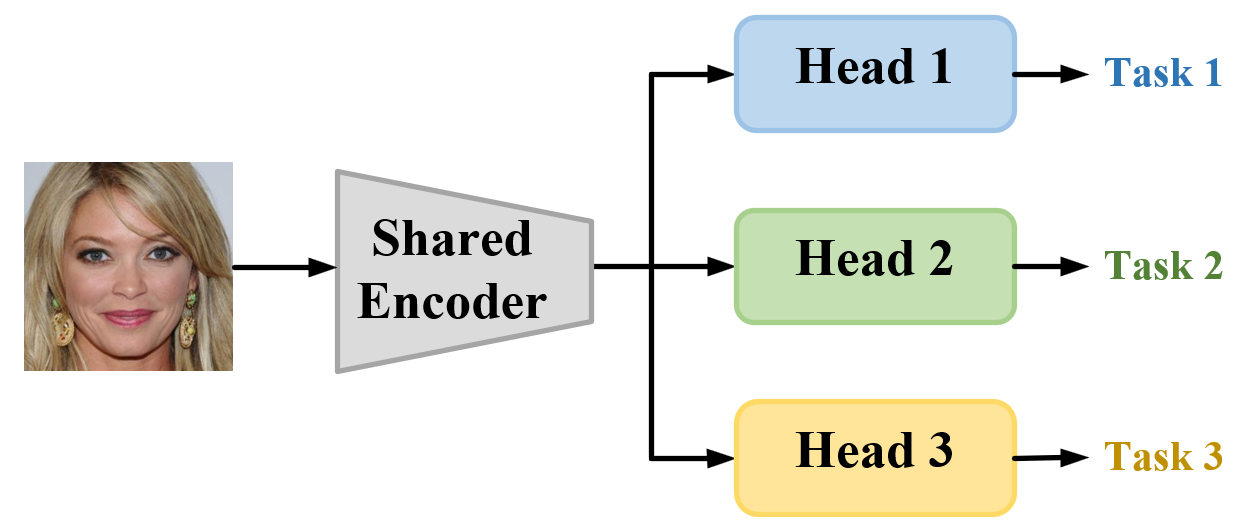}}
\vspace{5pt}
\subcaptionbox{The Proposed Task-adaptive Q-Face\label{fig:QF}}{\includegraphics[height=0.21\linewidth]{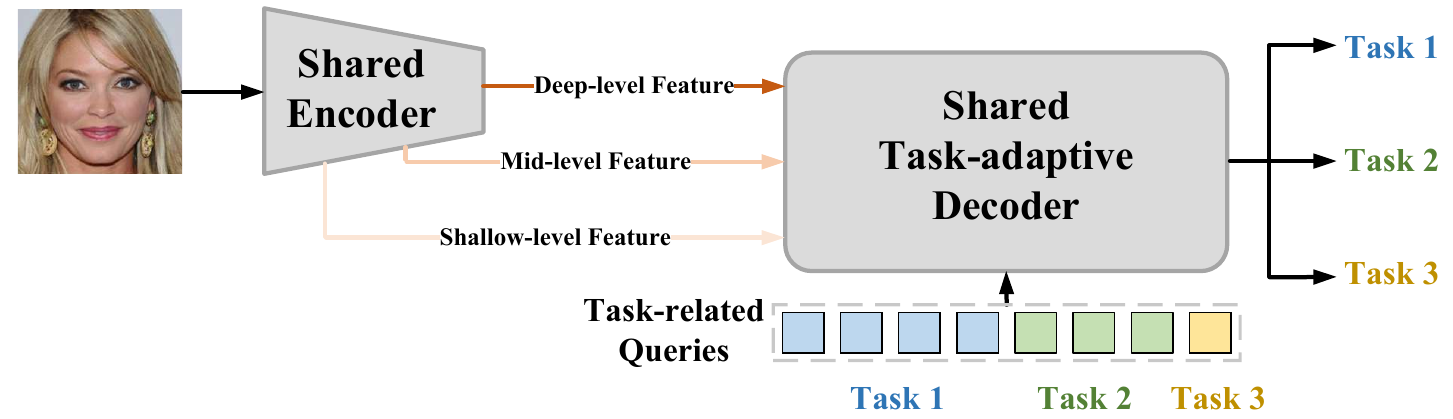}}
\vspace{-10pt}
 \caption{Compared to previous (a) task-specific methods and (b) multi-task methods, the proposed method can adaptively extract desired features from multi-stage feature maps according to the task requirements, thus exploiting the synergy between tasks.}
 \label{fig:compare}\vspace{-20pt}
\end{figure}

Inspired by this observation, some recent works \cite{zhang2016joint, ruiz2015emotions, wang2018weakly, han2017heterogeneous, ranjan2017hyperface} attempt to perform multiple face analysis tasks on a single face image and leverage the shared encoder and task-specific decoders to improve the performance of each task. However, different tasks often have different training objectives and constraints, and some objectives may even conflict with each other, e.g., face pose estimation and face expression recognition require pose-related and pose-independent features, respectively. Therefore, a multi-head model shown in Fig. \ref{fig:MH} cannot handle the trade-off between various tasks and can only share the encoder between similar tasks, such as face detection and face alignment \cite{zhang2016joint}, or face expression and action unit detection \cite{ruiz2015emotions, wang2018weakly}. To solve this problem, researchers have tried to classify face tasks into several categories and adaptively extract image features for each category \cite{ranjan2017all, qin2023swinface}. However, these methods require manual settings for different attribute categories, and it is almost infeasible to set up all face analysis tasks manually. In addition, those tasks in the same category still share the same features from the global pooling and lack explicit learning of task correlations, thereby failing to fundamentally resolve conflicts caused by inconsistent objectives of multiple face analysis tasks.

In recent years, the cross-attention mechanism in the decoder of \cite{vaswani2017attention} presents a possible choice to adaptively extract specific features and has been successfully validated in the area of object classification~\cite{liu2021query2label} and multi-modal representation learning~\cite{li2023blip}. In this paper, we propose a novel task-adaptive multi-task face analysis method, namely the Query-driven Face analysis method (Q-Face). Q-Face can simultaneously perform multiple face analysis tasks with a unified model and leverage the task correlations to enhance facial features. As shown in Fig. \ref{fig:QF}, our model consists of a shared encoder and a task-adaptive decoder. The shared encoder is pre-trained on a large-scale face dataset, which can extract general features for various tasks. The task-adaptive decoder first fuses the features from different layers of the shared encoder to maintain both local and global facial information and support multiple tasks. After that, our task adaptive decoder performs cross-attention between the fused multi-stage features and a set of query vectors that represent various tasks explicitly. In this way, our model can dynamically adjust the feature representation according to the task demand and avoid the interference of irrelevant features. We conduct extensive experiments on five public face data sets, namely CelebA \cite{liu2015deep}, RAF-DB \cite{li2017reliable}, EmotioNet \cite{fabian2016emotionet}, AgeDB \cite{AgeDB}, and BIWI \cite{fanelli2011real}. Our model achieves state-of-the-art performances on multiple tasks simultaneously. In addition, the visualization results demonstrate the superiority of our model in adaptively performing multiple face analysis tasks. In summary, this paper makes the following contributions:
\begin{itemize}
    \item We propose a unified and efficient framework for multi-task face analysis, which can simultaneously perform multiple tasks with a unified model and leverage the task correlations to enhance the face features.
    \item We design a task-adaptive decoder module that fuses the features from different layers and utilizes both local and global facial information to support multiple tasks. Besides, it can adaptively extract class-related features and avoid the feature dilution problem caused by global pooling. 
    \item We conduct extensive experiments on five public face data sets and achieve state-of-the-art performance on multiple face analysis tasks simultaneously. Our method opens up new possibilities for multi-task face analysis and shows the potential for both accuracy and efficiency.
\end{itemize}

\section{Related Works}
\label{sec:related}

\subsection{Face Analysis Tasks}

Face analysis includes a variety of tasks, such as face recognition, expression recognition, facial attribute recognition, and age estimation. In recent years, supervised-learning-based methods have achieved remarkable progress in most face analysis tasks by proposing novel loss functions and improving feature extraction models. For example, SphereFace\cite{liu2017sphereface}, CosFace \cite{wang2018cosface}, and ArcFace \cite{deng2019arcface} introduced angular margin-based loss functions to enhance the discriminative power of face features for face recognition. Meanwhile, researchers also explored local and global facial representations to capture the comprehensive facial information for expression recognition \cite{liu2022adaptive} and facial attribute recognition \cite{zhuang2018multi, hand2017attributes, mao2020deep} tasks. Regarding the age estimation tasks, several methods try to learn a label distribution and lead to a better understanding of the continuity of different labels \cite{tan2017efficient, zheng2022general, meanvarianceloss}. However, previous methods commonly treat the above-mentioned face analysis tasks as separate problems. Since the annotated data may not be available or sufficient for all tasks, the generalization ability of the learned representations is often limited. 

Recently, some researchers have focused on learning general face representations and transferring them to various face analysis tasks. Some tried to use multi-modal information associated with face images \cite{wiles2018self, li2019self, lu2020self, prajwal2020lip, sun2022masked}. Meanwhile, SSPL~\cite{shu2021learning}, MarLin~\cite{cai2022marlin}, and FaRL~\cite{zheng2022general} introduce a series of general facial representation learning frameworks to utilize the potential of the large-scale face dataset. Their learned representations can be transferred to several face analysis tasks. Despite the success of these general facial representation learning methods, they still need specific classification heads for each task and train them independently. To address this limitation, we design a task-adaptive module to extract features from different regions and depths adaptively according to different tasks. With the task-adaptive module, our model can both leverage the strong potential of the general face representation learning methods and mitigate conflicts between different face analysis tasks. As a result, our proposed method can perform various tasks more efficiently and simultaneously.

\subsection{Multi Task Learning}
Face analysis tasks are inherently interconnected and can benefit from synergy between different tasks. Consequently, some researchers try to develop multi-task face analysis models. These models aim to perform multiple face analysis tasks with a single model. Early multi-task face analysis methods focused on a limited number of related tasks. For example, MTCNN \cite{zhang2016joint} consists of three cascaded convolutional networks that can detect faces and landmark locations together. Meanwhile, several methods have been developed to perform face expression and action unit detection tasks jointly \cite{ruiz2015emotions, wang2018weakly}. However, these methods often rely on inter-task correlation between similar tasks, which limits their scalability to additional face analysis tasks.

To overcome this limitation, some researchers attempt to perform a more comprehensive set of face analysis tasks simultaneously. DMTL~\cite{han2017heterogeneous}, HyperFace \cite{ranjan2017hyperface}, and AIO \cite{ranjan2017all} employ a shared encoder to extract general features shared for various face analysis tasks. However, they still use multiple separate heads to generate task-specific outputs, which increases the parameters and computational cost in proportion to the number of face analysis tasks. Moreover, they often rely on human selection of features for each task, which may not always yield optimal results across different tasks. Recently, SwinFace \cite{qin2023swinface} attempts to select feature layers within the model dynamically. However, it still requires an artificial division of tasks, and most categories still share the same features obtained from the global pooling layer, which may exacerbate task conflicts. In this paper, we introduce a task-adaptive decoder, which adaptively selects the features required by various tasks. This approach enables the model to perform multiple face analysis tasks more efficiently simultaneously.
\section{Method}
\label{sec:method}

\begin{figure*}[t]
 \centering
 \includegraphics[width=0.95\linewidth]{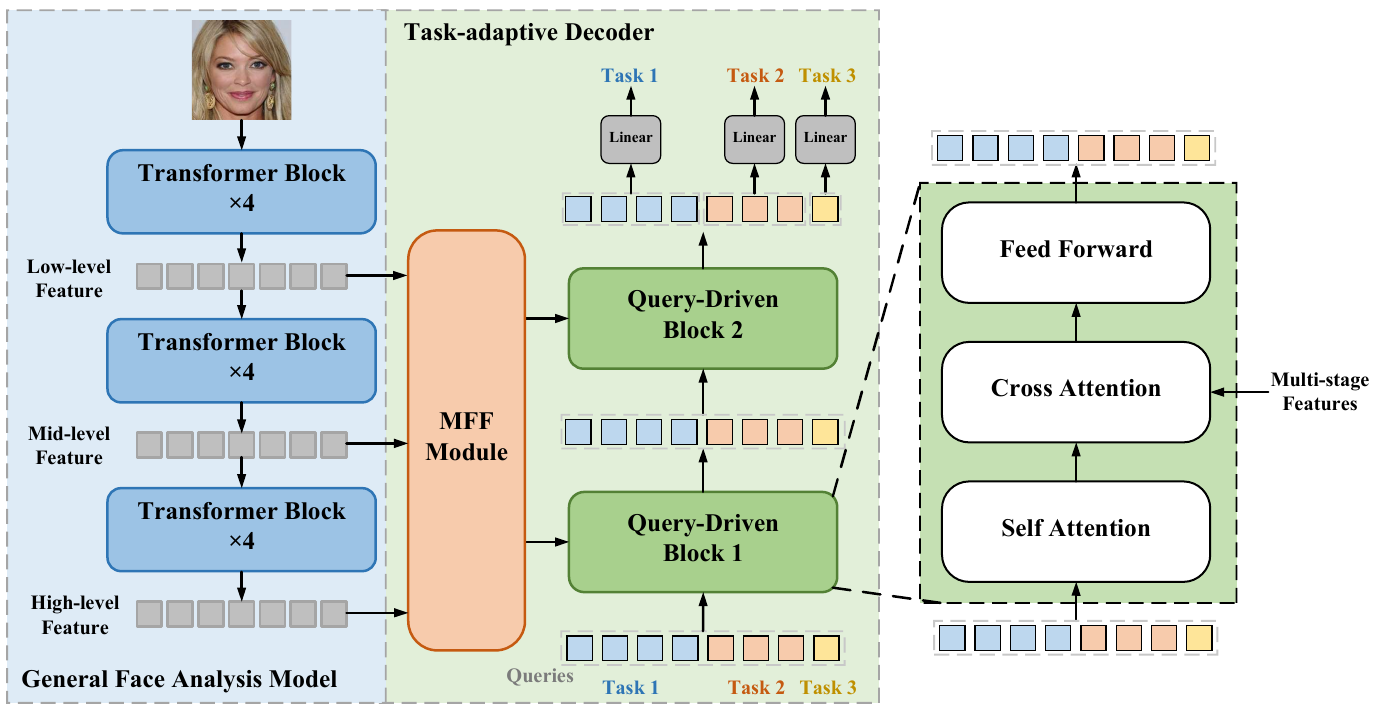}
 \caption{The framework for the proposed task-adaptive Q-Face. We design a task-adaptive decoder to extract task-related features adaptively from different stages and regions. This enables us to perform multiple face analysis tasks more effectively.}
 \label{fig:framework}\vspace{-15pt}
\end{figure*}

We propose a Query-driven Face Analyzer (Q-Face), a novel and efficient framework for multi-task face analysis. Q-Face can perform multiple face analysis tasks adaptively and reduce conflict between different tasks. The framework of the proposed Q-Face is shown in Fig.\ref{fig:framework}. Q-Face consists of a shared general face analysis model and a task-adaptive decoder. The general face analysis model is a pre-trained backbone that extracts facial features from the input image. To leverage the rich and diverse information in multi-stage features, the task-adaptive decoder first uses a multi-stage feature fusion (MFF) module to fuse the features from different stages. Then, a query-driven module is applied to adaptively perform various face analysis tasks by using the cross-attention blocks to query the fused multi-stage feature. The query-driven module enables the model to explicitly distinguish between different tasks and labels and leverage the correlation between tasks. In this section, we describe the details of each component and the training procedure.

\subsection{General Face Analysis Model}
Since our method aims to support multiple face analysis tasks, it is important to properly leverage a shared encoder with sufficient capacity. Here, we employ the Visual Transformer (ViT)~\cite{dosovitskiy2020image} structure. It split input image $\mathcal{I}$ into patches $\mathcal{P} = \{\mathcal{P}_1, \mathcal{P}_2, ..., \mathcal{P}_N\}$. Then, a fixed sine-cosine position embedding is added to each patch embedding to preserve the spatial information of the patches. The resulting embeddings, namely tokens, are concatenated with the [CLS] token and fed into the Transformer blocks $E_{ViT}$ for feature extraction. The output of $E_{ViT}$ is a feature map $ \mathcal{F}\in \mathbb{R}^{197\times 1024}$. 

To obtain a more suitable general embedding for face analysis tasks, we do not employ the commonly-used ImageNet~\cite{deng2009image} pre-training scheme of ViT. Instead, we pre-train the ViT model on a large dataset of face images using the mask image modeling (MIM) strategy~\cite{he2022masked}. During the pre-training process, $N_{m}$ patches are randomly masked from the input image, using a mask index set $\mathcal{M}\subset \{1,2,...N\}$. After the masking process, the remaining patches can be formulated as $\{\mathcal{P}_{i_1}, \mathcal{P}_{i_2}, ...\mathcal{P}_{i_{N_v}}\}$, where $N_v = N-N_m$ denotes the number of visible patches. Then, we use a two-layer transformer-based decoder $D_{MIM}$ to reconstruct the masked regions from the embeddings of the encoder $E_{ViT}$:
\begin{align}
   \{\mathcal{H}_{j_1}, \mathcal{H}_{j_2}, ..., \mathcal{H}_{j_{N_m}}\} = D_{MIM}(E_{ViT}(\mathcal{P}_{i_1}, \mathcal{P}_{i_2}, ..., \mathcal{P}_{i_{N_v}})). 
\end{align}
The objective function can be expressed as:
\begin{align}
    L_{MIM} = \sum_{i\in \mathcal{M}} L(\mathcal{H}_i, \mathcal{P}_i) = \sum_{i\in \mathcal{M}} \left\lVert \mathcal{H}_i- \mathcal{P}_i\right\rVert_1.
\end{align}
After pre-training, $E_{ViT}$ can learn both structural and textural contexts of face images. Its feature maps at different levels contain rich information about the face, which can be utilized for multiple face analysis tasks. To achieve better generalization performance, we utilize the feature maps of multiple layers as the shared feature for multiple face analysis tasks. We denote the feature map of the layer with index $i$ as $\mathcal{F}^i$, where $i\in \{1,2,...,12\}$.

\subsection{Multi-stage Feature Fusion Module}

\begin{figure*}[t]
 \centering
 \includegraphics[width=0.85\linewidth]{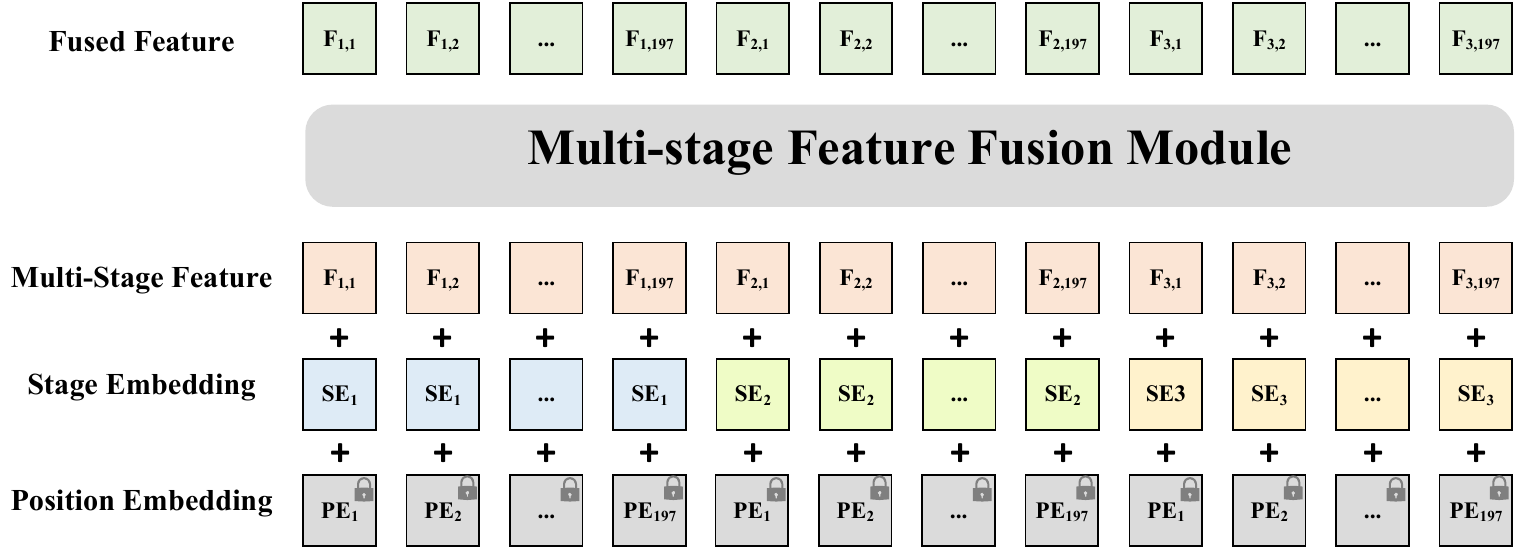}
 \caption{The diagram about the Multi-stage Feature Fusion (MFF) module. The proposed stage embeddings $SEs$ help our method to distinguish multi-stage features and enable the query-driven decoder to select task-related features more efficiently.}
 \label{fig:stageEMB}\vspace{-15pt}
\end{figure*}

We design the multi-stage feature fusion (MFF) module for two purposes. The first one is to fuse the features from different layers of the Transformer encoder so that the whole model can use both local and global facial information to support multiple tasks. The second one is to distinguish the features from different depths so that different tasks can select the features they rely on adaptively. 

We first concatenate the intermediate features $\mathcal{F}^4, \mathcal{F}^8\text{, and }\mathcal{F}^{12}$ of the Transformer encoder $E_{ViT}$ for the first purpose. To preserve the spatial information of the patches, we add a fixed sine-cosine position embedding $PE\in \mathbb{R}^{197\times 1024}$ to the concatenated feature map. To achieve our second purpose, we introduce a stage embedding $SE \in \mathbb{R}^{1024}$ to mark the feature map. Since we take the feature maps from three different depths, we set three $SEs$, namely $SE_1$, $SE_2$, and $SE_3$. Unlike the position embedding, the proposed stage embeddings are learnable vectors shared across the spatial dimension of $\mathcal{F}$. As shown in Fig.\ref{fig:stageEMB}, the stage embeddings $SEs$ are also added to the concatenated feature map. It allows the MFF module to distinguish between features from different layers and enables the query-driven decoder to select the class-related features more efficiently. We apply a transformer block to the concatenated feature map to merge the multi-stage features better. The output $F^*$ can be formulated as,
\begin{align}
    \mathcal{F}^* = Transformer(Concat(\mathcal{F}^4, \mathcal{F}^8, \mathcal{F}^{12})+SE+PE).
\end{align}
The Transformer block is a standard one in ViT, which consists of a multi-head self-attention and a feed-forward network. It captures the long-range dependencies and the semantic correlations among different patches and stages, and its output is ready for the following query-driven decoder module.

\subsection{Query-driven Decoder Module}

The query-driven decoder module aims to adaptively extract the desired features for each face analysis task. To achieve this task flexibility, we need to mark each task using a proper task query and then design a mechanism to select the related features automatically. We notice that the query-based decoder in Query2Label \cite{liu2021query2label} explores the idea of solving multi-label classification via query-based cross-attention. Inspired by it, we further improve the design of the query-based module and enable it to solve multi-task problems. We carefully design a query-driven decoder (QD) module that adaptively integrates features from different regions and levels according to different tasks and labels.

For each face analysis task, we mark it with a set of learnable query vectors. Each vector has a dimension of 1024 and represents a category label. We denote the query vectors as $\mathcal{Q}\in \mathbb{R}^{N\times 1024}$, where $N$ is the number of labels for a given task. Compared to the Query2Label method, our method combines labels from different face analysis tasks, making exploiting correlations between different tasks possible. In addition, we try to exploit query vectors' potential by using them for both classification and regression tasks, which allows the model to have stronger generalization capabilities. Then, we use these query vectors to select the multi-stage feature map $\mathcal{F}^*$ with two transformer blocks. As shown in the right of Table~\ref{fig:framework}, each transformer block consists of a multi-head self-attention layer, a multi-head cross-attention layer, and a feed-forward network. The transformer block $i$ updates the queries $\mathcal{Q}_{i-1}$ as follows:
\begin{align}
    \text{self-atten:} &&&Q_i^1 = MultiHead(\hat{\mathcal{Q}}_{i-1}, \hat{\mathcal{Q}}_{i-1}, \mathcal{Q}_{i-1}), \\
    \text{cross-atten:} &&&\mathcal{Q}_i^2 = MultiHead(\hat{\mathcal{Q}}_{i}^1, \hat{\mathcal{F}}, \mathcal{F}),\\
    \text{FFN:} &&&\mathcal{Q}_i = FFN(\mathcal{Q}_i^2),
\end{align}
where the $\hat{*}$ means the vectors modified by adding the position embedding. $\mathcal{Q}_i^1$ and $\mathcal{Q}_i^2$ are two intermediate variables. The $MultiHead(query, key, value)$ and $FFN(x)$ functions are the same as defined in the standard Transformer decoder \cite{vaswani2017attention}. The multi-head self-attention layer allows the model to capture the dependencies and correlations among the labels. Meanwhile, the cross-attention layer allows the model to extract desired task-specific features from different regions of the feature map adaptively. 

The output of the query-driven decoder is a set of feature vectors corresponding to each label. After that, we map those feature vectors to a set of logits for face analysis tasks:
\begin{align}
    z = f(QD(\mathcal{Q}, \hat{\mathcal{F}^*}, \mathcal{F}^*)).
\end{align}
$QD$ denotes the query-driven module, which takes three inputs: the query, key, and value. $f$ denotes the linear function, while the $z\in \mathbb{R}^N$ denotes the logits output of our model. Each task has its specific loss function, which is elaborately selected based on the task type and the data distribution. For example, we use the cross-entropy loss for face expression recognition and the binary-cross-entropy loss for the action unit detection and attribute classification tasks.

\subsection{Training Procedure}

After pre-training the general face analysis model with a large-scale face dataset, we fine-tune our model simultaneously with multiple face analysis tasks, including face expression recognition, action unit detection, face attribute analysis, age estimation, and face pose estimation. To better distinguish, we represent each face image and its annotations as ($\mathcal{I}_i$,$y_i$,$t_i$), where $\mathcal{I}_i$ is the $i$-th image, $y_i$ is the ground truth of the image, and $t_i$ is the task that the image belongs to. To prevent the model from focusing too much on specific tasks during alternation, we sample the batch of each task simultaneously and accumulate their loss functions together. The overall loss functions can be formulated as follows,
\begin{equation}
L = \sum_{i=1}^N \alpha_{t_i}L_{t_i}(y_i, z_{i}^{t_i}),
\end{equation}
where N is the number of images, $L_{t_i}$ is the loss function for the $t_i$-th task. $z_{i}^{t_i}$ denotes the predicted label vectors related to the task $t_i$ and the $i$-th image. $\alpha_{t_i}$ is the weight of the task $t_i$. This training procedure leads to a more balanced and efficient model to handle various face analysis tasks.

\section{Experiments}
\label{sec:exp}
To evaluate our method on multiple face analysis tasks, we employ five challenging datasets: RAF-DB~\cite{li2017reliable} for evaluating face expression classification, EmotioNet \cite{fabian2016emotionet} for evaluating the action unit detection task, CelebA \cite{liu2015deep} for evaluating the face attribute recognition task, AgeDB \cite{AgeDB} for evaluating the age estimation and gender classification task, and BIWI \cite{fanelli2011real} for evaluating the facial pose estimation tasks. We compare our multi-task Q-Face with state-of-the-art methods optimized for each specific task. To demonstrate the effectiveness of the task-adaptive decoder, we also set up a straightforward multi-task model using the same encoder as Q-Face but with a conventional multi-head decoder.


\subsection{Implementation Details}

\textbf{Pre-training of the general face analysis.} We employ the ViT-B/16~\cite{dosovitskiy2020image} as our image encoder. It is a 12-layer visual Transformer with 768 hidden units. The input image is with the size of 224$\times$224, and is sliced into 14$\times$14 patches. To pre-train the ViT-B/16, we utilize the large-scale dataset Glint-360k~\cite{An2021PartialFT}, which contains 17 million face images. Our training setup involves 4 Nvidia A100 GPUs with a batch size of 2048. We use the AdamW optimizer with a weight decay of 0.05 and a cosine learning rate schedule that warms up to 1e-3 in the first five epochs and then decreases to 1e-6 in the subsequent 75 epochs.

\noindent\textbf{Finetuning with the task-adaptive decoder.} When fine-tuning our model on multiple tasks, we use the following settings: batch size=512, layer decay=0.85, drop path rate=0.2, and weight decay=0.05. The embedding dimension of the query-driven decoder is 1024. We fine-tune the pre-trained model for 100 epochs using an AdamW optimizer and a cosine learning rate schedule.

\begin{table*}[t]
 \centering
  \caption{Comparing with SOTA method in two tasks, including (a) face expression classification on RAF-DB, and (b) attribute recognition on CelebA.}
 \vspace{-5pt}
 \begin{subtable}{0.47\linewidth}
 \centering
 \vspace{-5pt}
 \begin{tabular}{ccc}
 \toprule
 Method & Multi-task & Accuarcy\\
 \midrule
 SCN \cite{wang2020suppressing}& & 87.03\\
 EfficientFace \cite{zhao2021robust}& & 88.36\\
AMP-Net \cite{liu2022adaptive} & & 89.25 \\
TransFER \cite{xue2021transfer} & & 90.91 \\
 APVIT \cite{xue2022vision}& & 91.98\\
 POSTER++ \cite{mao2023poster}& & 92.21\\
 ARBEx \cite{wasi2023arbex}& & 92.47\\
 SwinFace \cite{qin2023swinface} &\checkmark &90.97 \\
 \midrule
 Multi Head &\checkmark &92.05\\
 Q-Face &\checkmark& 92.86\\
 \bottomrule
 \end{tabular}
 \caption{Performance on RAF-DB}
 \label{tab:RAF}
 \vspace{-10pt}
 \end{subtable}
 \begin{subtable}{0.47\linewidth}
 \centering
 \vspace{-5pt}
 \begin{tabular}{ccc}
 \toprule
 Method & Multi-task  & Accuarcy\\ 
 \midrule
PANDA-1 \cite{zhang2014panda} && 85.43\\
LNets+ANet \cite{liu2015deep} && 87.33\\
MOON \cite{rudd2016moon} & &90.94\\
NSA \cite{mahbub2018segment} && 90.61\\
MCNN-AUX~\cite{hand2017attributes} && 91.29\\
DMM-CNN \cite{mao2020deep} && 91.70\\
MCFA \cite{zhuang2018multi}  &\checkmark& 91.23\\
SwinFace \cite{qin2023swinface} &\checkmark& 91.32\\
 \midrule
 Multi Head&\checkmark &91.25\\
 Q-Face &\checkmark &91.56\\
 \bottomrule
 \end{tabular}
 \caption{Performance on CelebA}
 \label{tab:CelebA}
 \vspace{-10pt}
 \end{subtable}
 \vspace{-15pt}
\end{table*}

\subsection{Face Expression Classification} 
To evaluate the face expression classification performance of our Q-Face, we employ the RAF-DB dataset \cite{li2017reliable}, a widely used facial expression database. The dataset contains 15,339 face images, divided into 12,271 training images and 3,068 test images. Each image is labeled with one of the seven basic expressions: surprise, fear, disgust, happiness, sadness, anger, and neutral. For this task, we design a 7-dimensional learnable vector to select expression-related features through the query-drive module. Meanwhile, we use the cross-entropy loss as the fine-tuning loss and top-1 accuracy as the evaluation metric. 

We compare the proposed method with both the state-of-the-art methods designed specifically for expression classification task \cite{xue2022vision, mao2023poster, wasi2023arbex}, and the multi-task method SwinFace \cite{qin2023swinface}, which can perform face detection, alignment, recognition, and expression classification simultaneously. As shown in Table \ref{tab:RAF}, our method outperforms the state-of-the-art methods by a large margin, achieving 92.86\% top-1 accuracy on the test set. This demonstrates that Q-Face can accurately classify facial expressions in real-life scenes and utilize the shared information between different facial analysis tasks. In addition, the significant improvement of our method over the multi-head method and SwinFace also reflects our method's ability to perform multiple face analysis tasks simultaneously and effectively.

\subsection{Face Attribute Recognition} 

Face attribute recognition aims to predict semantic attributes of face images such as gemder, hair color, glasses, beard, and smile. Compared with the expression recognition task, the face attribute recognition task contains a richer set of attributes and requires a model with a deeper understanding of the face images' local and global features. We utilize the CelebA~\cite{liu2015deep} dataset to evaluate the face attribute recognition task. CelebA dataset contains more than 202K face images with 40 attribute annotations per image. Following \cite{shu2021learning, zheng2022general}, we use 162,770 images for training and 19,962 images for testing. Our method uses the binary cross-entropy (BCE) loss as the loss function and the average accuracy as the evaluation metric. Table \ref{tab:CelebA} compares our method with state-of-the-art attribute recognition and multi-task methods. The performance of our method is comparable to that of SOTA methods, only behind DMM-CNN\cite{mao2020deep} that uses human prior and classifies face attributes into different groups. Our model also outperforms existing multi-task face analysis methods~\cite{zhuang2018multi,qin2023swinface}. These results demonstrate that our task-adaptive decoder can perform multiple face analysis tasks simultaneously without human prior. Since recognizing different attributes requires diverse facial features, these results also show the task-adaptive decoder can support multiple tasks by fusing local and global facial information.

\subsection{Action Unit Detection} 

\begin{table*}[t]
 \centering
  \caption{Comparing with other facial action unit detection methods on EmotioNet dataset. The average F1-score (\%) of 12 AU classes is used for comparison.}
 \vspace{-5pt}
 \begin{tabular}{ccccccccccccccccc}
 \hline
 AU & 1&2&4&5&6&9&12&17&20&25&26&43&Avg.\\
 \hline
 ResNet-34~\cite{he2016deep}& 55.6& 41.1& 70.1& 46.6& 80.2& 59.2& 90.7& 45.5& 45.1& 94.2& 58.7& 60.5& 62.3 \\
 MLCT~\cite{xing2018multi}& 57.8& 44.8& 73.7& 50.1& 82.8& 58.1& 91.8& 44.8& 37.1& 95.1& 61.6& 63.4& 63.4 \\
 Mean-teacher~\cite{tarvainen2017weight}& 55.5& 46.3& 71.1& 48.6& 81.6& 61.7& 91.0& 46.7& 43.5& 94.7& 60.2& 63.9& 63.7 \\
 Co-training~\cite{zhao2015joint}& 58.3& 48.4& 70.0& 50.4& 83.1& 64.4& 91.7& 49.9& 47.1& 95.0& 60.0& 66.9& 65.5 \\
 MLCR~\cite{niu2019multi} &  61.4& 49.3 &  75.9&54.1 &  83.5 &  68.3& 92.0& 50.8& 53.5 &  95.2& 65.1& 68.1& 68.1 \\
 \hline
 Multi Head & 65.2 & 41.4 & 78.3 & 59.6 & 83.5 & 69.5 & 91.9 & 58.5 & 57.6 & 96.3 & 61.0 & 71.2 & 69.5 \\
 Q-Face & 64.0 & 43.7 & 78.7 & 55.6 & 83.3 & 71.6 & 91.7 & 63.7 & 54.0 & 96.5 & 67.4 & 74.3 & 70.4 \\
 \hline
 \end{tabular}
\label{tab:EmotioNet}
\vspace{-15pt}
\end{table*}

Action unit detection is a multi-label classification task for specific action units (AUs). We evaluate our method on EmotioNet dataset \cite{fabian2016emotionet}, a database with 20,722 images. Each image was manually labeled with 12 AUs by experts. Following \cite{niu2019multi}, we randomly select 15,000 images for training and use the remaining images for testing. The AU detection task focuses more on local facial information and requires the model to deeply understand different face components. 12 learnable query vectors are used for the adaptive feature extraction. We supervise them using a BCE loss function and use the F1 score as the evaluation metric. Table \ref{tab:EmotioNet} shows the comparison results. We can see that the model's performance is significantly improved after introducing the task-adaptive decoder. This result reflects that the proposed Q-Face model can better adaptively focus on different facial regions according to different labels. This ability improves the model's robustness in dealing with fine-grained, face region-related analysis tasks. Meanwhile, our model also outperforms the multi-head model. This improvement suggests that by introducing a task-adaptive decoder, our proposed Q-Face model can effectively mitigate the performance loss caused by multi-task conflicts and help the model better adapt to multiple tasks.

\begin{table*}[t]
 \centering
  \caption{Comparing with SOTA method in age estimation task on AgeDB dataset. The accuracy (\%) of gender classification is also reported for evaluation.}
 \vspace{-5pt}
 \centering
 \begin{tabular}{cccc}
 \toprule
 Method& Multi-task & Age MAE & Gender Acc.\\
 \midrule
 AgeED~\cite{tan2017efficient} && 6.22 & -\\
 FaRL \cite{zheng2022general} && 5.64 & \\
 MV~\cite{meanvarianceloss}&&5.31&-\\
 CMT~\cite{conf68} &\checkmark& 6.19 & 94.0 \\
 Xie et al.~\cite{conf69} &\checkmark& 5.74 & 97.1 \\
 DCDL~\cite{sun2021deep} &\checkmark& 5.27 & 97.1 \\
 \midrule
 Multi Head &\checkmark& 4.82 & 97.9\\
 Q-Face &\checkmark& 4.67 & 98.1\\
 \bottomrule
 \end{tabular}
 \label{tab:AgeDB}
 \vspace{-15pt}
 \end{table*}
\vspace{-5pt}
\subsection{Age Estimation} 

Age estimation aims to predict a person's age from the input images. We evaluated our method on the AgeDB dataset \cite{AgeDB}, which contains 16,488 images. Each face is labeled with identity, age, and gender. We use 80\% of the data for training and the rest for testing. Unlike the previous three datasets, the AgeDB dataset can evaluate the model's performance in both classification and regression tasks. Here, we apply three learnable query vectors to pool features related to age, male and female. We use smoothed-L1 loss as the loss function and mean absolute error as the evaluation metric. To help the model understand the continuity of different labels, we add random Gaussian noise with a standard deviation of 1 as a perturbation to the corresponding age labels during training. This technique improves the robustness and generalization of the model and avoids over-fitting to discrete labels. Table \ref{tab:AgeDB} compares our method with SOTA age estimation methods. Although \cite{conf68}, \cite{conf69}, and \cite{sun2021deep} try to enhance the age estimation performance with the gender classification task, they still cannot transfer their models to other face analysis tasks. With the task-adaptive decoder, Table \ref{tab:AgeDB} shows that our model can perform both classification and regression tasks better. Meanwhile, compared to the multi-head model, our model can still achieve a better age estimation result while better ensuring the performance of other tasks. This reflects that our task-adaptive decoder can effectively avoid the feature dilution problem caused by global pooling.
\begin{table*}[t]
     \centering
  \caption{Comparing with SOTA method in pose estimation task on BIWI dataset. The mean average error of three rotation angles, yaw, pitch, and roll, are evaluated.}
 \vspace{-5pt}
 \begin{tabular}{cccccc}
 \toprule
 Method & Multi-task&Yaw &Pitch & Roll & MAE\\
 \midrule
HopeNet~\cite{ruiz2018fine} && 3.29 & 3.39 & 3.00 & 3.23\\
FSA-Net \cite{yang2019fsa} && 2.89 & 4.29 & 3.60 & 3.60\\
TriNet \cite{cao2021vector}&& 2.93 & 3.04 & 2.44 & 2.80\\
FDN \cite{zhang2020fdn}& &3.00 & 3.98 & 2.88 & 3.29\\
6DRepNet \cite{hempel20226d} && 2.69 & 2.92 & 2.36 & 2.66\\
 \midrule
 Multi Head &\checkmark& 3.37& 2.64 & 2.86 & 2.95\\
 Q-Face &\checkmark& 3.17 & 3.00 & 2.58 & 2.92\\
 \bottomrule
 \end{tabular}
 \label{tab:BIWI}
 \vspace{-15pt}
\end{table*}

\subsection{Pose Estimation} 

Our model can also handle the pose estimation task, which aims to estimate the orientation of a face relative to the camera in 3D space. We evaluate our method on the BIWI dataset \cite{fanelli2011real}, which contains 15678 images of 20 subjects captured by the Kinect sensor. Following \cite{yang2019fsa, hempel20226d}, we employ the commonly used 70-30 protocol, where a 70\% subset is randomly selected for training. Compared with previous tasks, the pose estimation task requires the model to have a global view of the face images. Referring to \cite{hempel20226d}, we use six labels for the task-adaptive decoder, corresponding to the 6D representation of the head pose. We use the geodesic loss as the loss function and the mean average error of the Euler angles as the evaluation metric. To improve the labels' continuity, we add a Gaussian perturbation with a standard deviation of 0.01 in the rotation matrix. We only use the perturbation during training and use the original labels for testing. Table \ref{tab:BIWI} compares our method and other SOTA methods. It can be seen that our model performs comparably with these methods. It is worth noting that this is reached with a guaranteed low impact on the performance of other face analysis tasks, further demonstrating our model's adaptive ability in multiple tasks.

\vspace{-5pt}
\subsection{Visualization}

\begin{figure*}[t]
 \centering
 \includegraphics[width=0.9\linewidth]{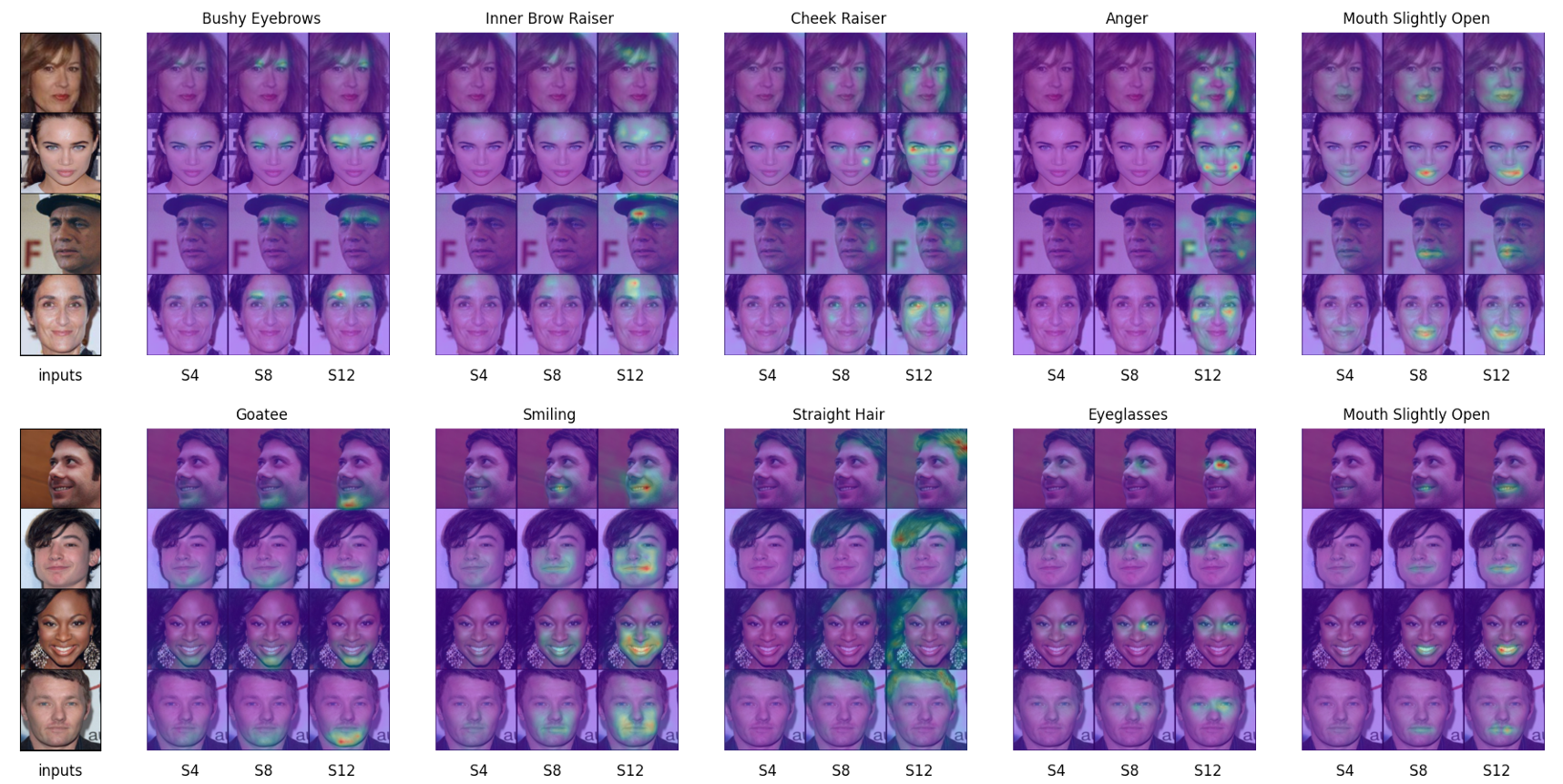}
 \vspace{-10pt}
 \caption{Visualization of the attention map from the query-driven module. S4, S8, and S12 represent the attention map related to different stages of features. Q-Face can focus on the relevant regions adaptively. For instance, it pays more attention to the eye region for labels such as eyeglasses and bushy eyebrows.}
 \label{fig:QDM-DiffReg}
 \vspace{-10pt}
\end{figure*}

\begin{figure*}[t]
 \centering
 \includegraphics[width=0.7\linewidth]{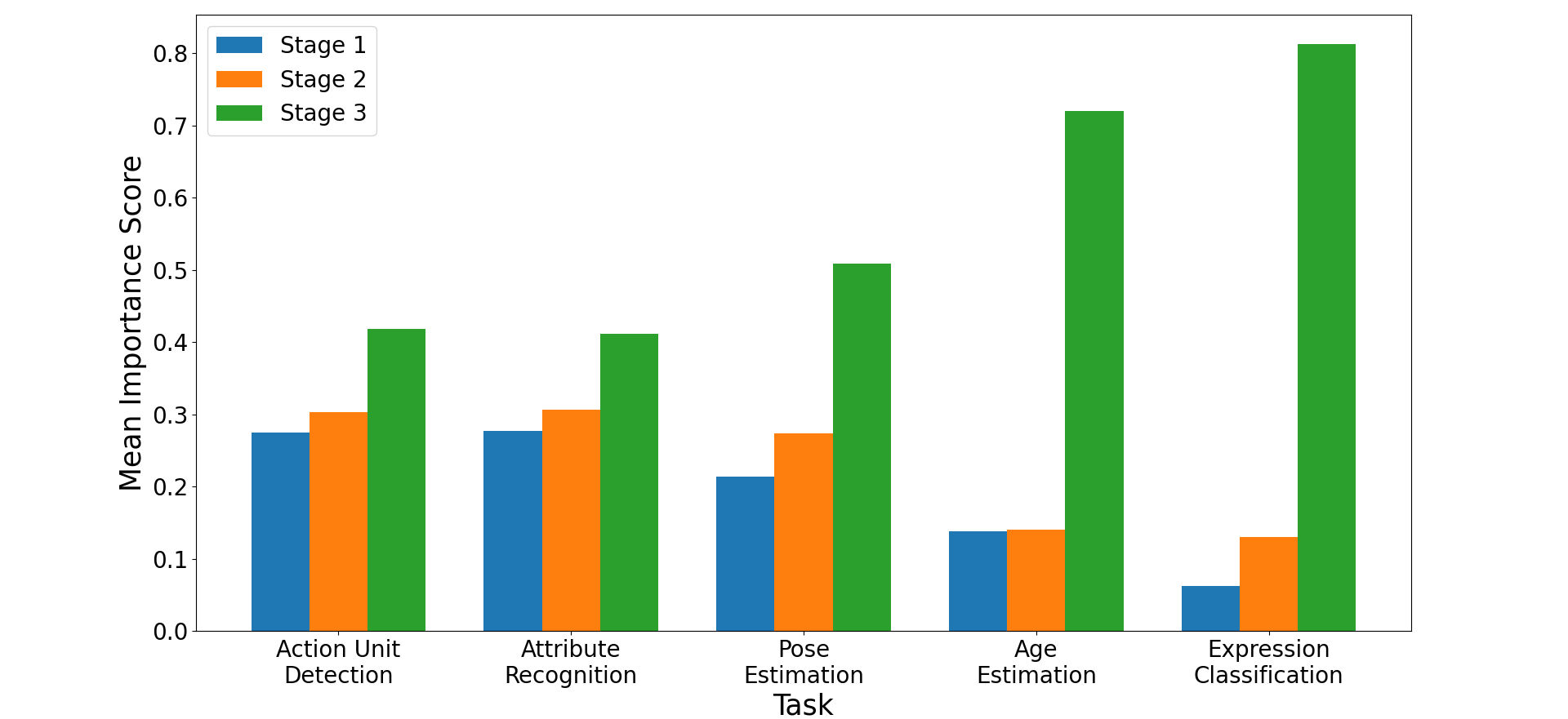}
 \vspace{-10pt}
 \caption{Visualization of how different tasks rely on different stages of features. Our model adaptively selects deeper features for the expression recognition task and age estimation task. In contrast, our model prefers to use shallow features for other tasks.}
 \vspace{-15pt}
 \label{fig:QDM-DiffStage}
\end{figure*}

The cross-attention layer in the query-driven module can show how much attention our model pays to each region and layer of face images. To verify the important role of the QD module, we visualize the cross-attention maps for each category in the first transformer block. The visualization results are shown in Fig. \ref{fig:QDM-DiffReg}. We can see that Q-Face can distinguish different labels and focus on the relevant regions adaptively. For instance, our model pays more attention to the eye region for labels such as eyeglasses and bushy eyebrows. In contrast, the mouth region received more attention when performing smile detection and other mouth-related tasks. This demonstrates the QD module's ability to select the appropriate regions for different tasks and labels, thereby minimizing task interference and effectively performing multiple face analysis tasks.

We also analyze how different tasks rely on different stages of the feature map. Fig \ref{fig:QDM-DiffStage} displays the visualization results. We can find significant differences in how much the model focuses on local and global features in different tasks. Specifically, our model relies more on deeper features with better global consistency for expression recognition task and age estimation task. On the other hand, our model prefers to use shallow features with better local and texture features for tasks such as action unit recognition, attribute analysis, and pose estimation. This visualization indicates our model's ability to adaptively select the desired feature for different tasks without any manual intervention. This also shows the importance of the multi-level feature fusion module, which helps the model to distinguish and utilize the features from different stages.

\vspace{-5pt}
\subsection{Ablation Studies}

To evaluate the effectiveness of each component in Q-Face, we compare our method with two baselines: a straightforward multi-head model and a group of task-specific models separately trained for each task. We also ablate the multi-stage feature fusion (MFF) module to examine its effect. Inspired by the ABAW multi-task learning (MTL) competition \cite{kollias2022abaw, kollias2023abaw}, we adopt the average performance metric (AVG) to measure the synergy between tasks. We use the average F1 score for each category in the classification task and the average covariance correlation coefficient (CCC) in the regression task to evaluate model performance. Both metrics range from 0 to 1, with higher values indicating better predictions. Table \ref{tab:Ablation} shows that our model outperforms the multi-head model in all face analysis tasks, demonstrating its robustness and superiority in handling multiple tasks. The multi-head model, on the other hand, suffers from task conflicts caused by the global pooling. By integrating the MFF and query-driven modules, the proposed Q-Face can select features adaptively from different stages and regions for different tasks, resulting in simultaneous and efficient face analysis. The MFF module significantly improves the performance of action unit detection, attribute recognition, and pose estimation tasks. These improvements show the importance of utilizing multi-stage features and have further verified the observations made in the visualization experiments. Our model also outperforms the task-specific model in most tasks, suggesting that a task-adaptive decoder can exploit the inter-task correlation to enhance face analysis.

\begin{table}[t]
 \centering
 \caption{Ablation Study of Multi-task learning strategy. "MT", "QD", and "MFF" denote whether the model contains the multi-task, the query-driven module, and the multi-stage feature fusion module, respectively.}
 \label{tab:Ablation}
 \vspace{-3pt}
 \begin{tabular}{l | c c c | c c c | c c | c c c}
 \toprule
 \multirow{2}{*}{Method} & \multirow{2}{*}{MT} & \multirow{2}{*}{QD} & \multirow{2}{*}{MFF} & \multicolumn{3}{c|}{F1-Score} & \multicolumn{2}{c|}{CCC} & \multirow{2}{*}{AVG}\\
 \cline{5-9}
 &&&& CelebA& EmotioNet & RAF-DB & AgeDB & BIWI & \\
 \midrule
 Task Specific Model& &&& 72.12 & 70.86 & 88.90 & 93.48 & 94.39 & 83.95\\
  Multi-head Model& \checkmark &&& 71.20 & 69.49 & 86.86 & 92.84 & 95.58 & 83.19\\
 Q-Face (w/o MFF)& \checkmark &\checkmark&& 72.79 & 69.04 & 87.67 & 93.23 & 95.78 & 83.70\\
 Q-Face & \checkmark &\checkmark&\checkmark& 73.58 & 70.36 & 87.96 & 93.53 & 96.41 & 84.37\\
 \bottomrule
 \end{tabular}
 \vspace{-13pt}
\end{table}

\vspace{-5pt}
\section{Conclusion}
To improve the multi-task face analysis task, we propose a novel task-adaptive multi-task face analysis model, which can perform multiple face analysis tasks simultaneously with a unified model and leverage the task correlations to enhance the face features. Our study shows that a query-driven module can help the model extract desired features from the different regions and levels adaptively, thus reducing conflicts between different tasks and leveraging the synergy among related tasks. In addition, the task-adaptive decoder can fuse the features from different layers and utilize both local and global facial information to support multiple tasks. With the proposed task-adaptive module, our model can perform multiple face analysis tasks simultaneously and more efficiently.

\noindent\textbf{Limitation.} Although applying the task-adaptive decoder can effectively mitigate conflicts between tasks and thus improve the model’s performance on multiple face analysis tasks, the method proposed in this paper still struggles to complete a new face analysis task and classify an unlabelled category. In the future, we will explore how the pre-trained natural language processing models can be used to establish the correlation between known and unknown tasks and improve the model’s performance in the open-set face analysis tasks.

%
%
\bibliographystyle{splncs04}
\bibliography{egbib}
\end{document}